\documentclass[letterpaper, 10 pt, conference]{ieeeconf}

\IEEEoverridecommandlockouts
\usepackage{cite}
\usepackage{amsmath,amssymb,amsfonts}
\usepackage{algorithmic}
\usepackage{graphicx}
\usepackage{textcomp}
\usepackage{xcolor}
\usepackage{algorithm}
\usepackage{url}
\usepackage{tabularx}
\usepackage{booktabs}
\usepackage{makecell}
\usepackage{float}
\usepackage{multirow} 
\usepackage{caption}
\usepackage{subcaption}
\usepackage{pifont}

\usepackage{placeins}

\usepackage{arydshln}

\definecolor{node}{RGB}{106, 168, 79} 
\definecolor{edge}{RGB}{60, 120, 216} 
\newcommand{\node}[1]{{\color{node}{#1}}}
\newcommand{\edge}[1]{{\color{edge}{#1}}}

\definecolor{navy}{RGB}{0,0,128}

\def\BibTeX{{\rm B\kern-.05em{\sc i\kern-.025em b}\kern-.08em
    T\kern-.1667em\lower.7ex\hbox{E}\kern-.125emX}}
\begin{document}

\title{
\textbf{
Towards Robotic Tree Manipulation: Leveraging Graph Representations}\\
\thanks{C. H. Kim, M. Lee, O. Kroemer and G. Kantor are with Carnegie Mellon University, Pittsburgh PA, USA \texttt{\{chunghek, moonyoul, okroemer, kantor\}@andrew.cmu.edu}}
\thanks{This work was supported in part by NSF Robust Intelligence 1956163, and NSF/USDA-NIFA AIIRA AI Research Institute 2021-67021-35329.}
}

\author{Chung Hee Kim, Moonyoung Lee, Oliver Kroemer, George Kantor}

\maketitle

\begin{abstract}

There is growing interest in automating agricultural tasks that require intricate and precise interaction with specialty crops, such as trees and vines. However, developing robotic solutions for crop manipulation remains a difficult challenge due to complexities involved in modeling their deformable behavior. In this study, we present a framework for learning the deformation behavior of tree-like crops under contact interaction. Our proposed method involves encoding the state of a spring-damper modeled tree crop as a graph. This representation allows us to employ graph networks to learn both a forward model for predicting resulting deformations, and a contact policy for inferring actions to manipulate tree crops. We conduct a comprehensive set of experiments in a simulated environment and demonstrate generalizability of our method on previously unseen trees. Videos can be found on the project website: \textcolor{navy}{\url{https://kantor-lab.github.io/tree_gnn}}

\end{abstract}


\section{Introduction}

\begin{figure}[t]
\centering 
\includegraphics[width=\columnwidth]{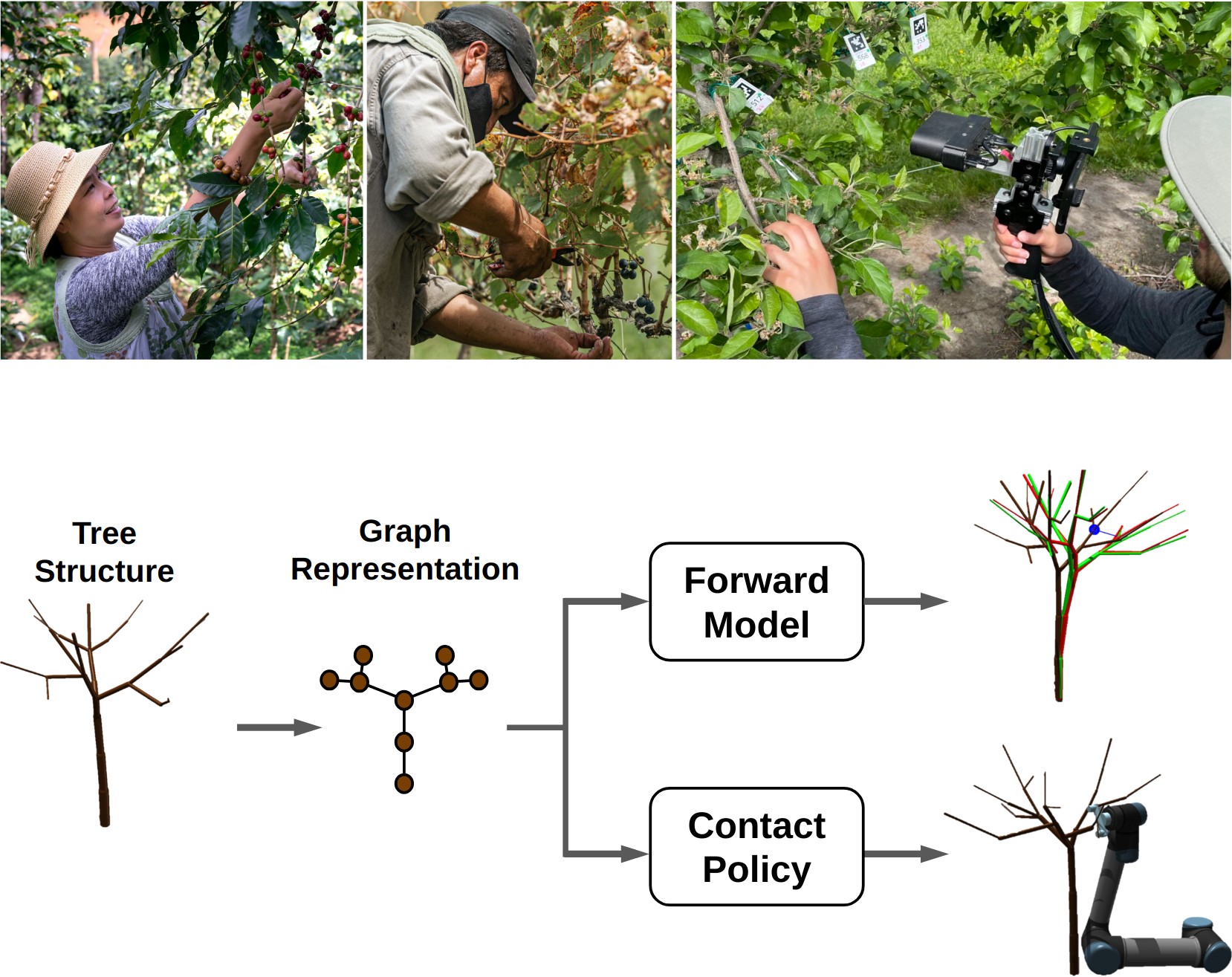}
\setlength{\unitlength}{1cm}
\begin{picture}(0,0)
\put(-3.1, 4.45){\footnotesize (a)} 
\put(-0.75,  4.45){\footnotesize (b)} 
\put(2.3, 4.45){\footnotesize (c)} 
\put(0.0,  0.6){\footnotesize (d)} 
\end{picture}
\vspace{-13pt}
\caption{Laborious agricultural tasks often require delicate contact interaction with crops. For example, a worker is (a) harvesting, (b) pruning, and (c) inspecting \cite{freeman2022autonomous} crops with their right hand, while pulling/pushing on branches with their left hand. (d) Overview of our proposed framework for learning the deformation behavior of tree-like crops under contact interaction.}
\label{fig:intro} 
\vspace{-13pt}
\end{figure}
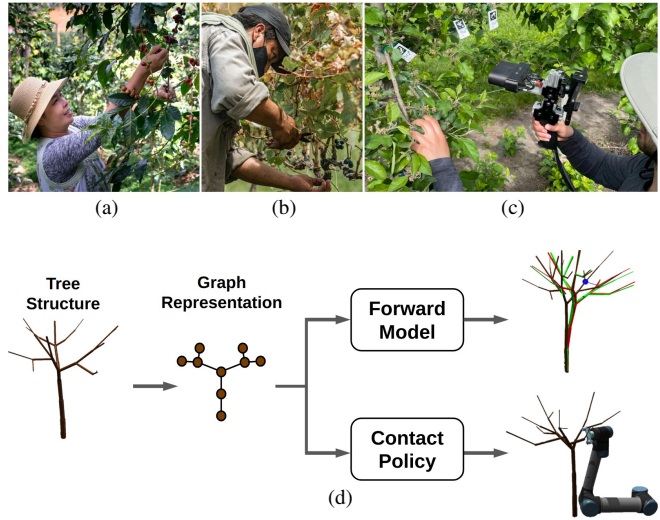

In response to labor shortages in agriculture, there is growing interest in adopting labor-saving mechanization, notably robotics, for automating agricultural tasks \cite{laborshortage}. Many of these tasks such as harvesting, pruning, and inspecting involve contact interactions like pushing or pulling on branches to reveal obstructed objects as illustrated in Fig \ref{fig:intro}. However, the deformable nature of tree branches presents a significant challenge for robot manipulation. Deformable objects require characterization in a higher-dimensional configuration space to accurately represent their states. Consequently, modeling the geometry and dynamics of branches becomes a non-trivial task  \cite{deformable_plant_model}. Even more challenging is the task of selecting actions for manipulating trees, such as determining optimal contact points and the appropriate direction for perturbation. 

One approach to modeling tree branch deformation under applied force is to use Finite Element Analysis (FEA) \cite{FEA_tree}. While FEA allows precise offline analysis, real-time deployment poses difficulties and demands an accurate 3D model beforehand. Alternatively, we use a method that approximates tree motions through kinematics and dynamics models. Past studies have constructed geometric models of trees with rigid links articulated by spring-damper joints \cite{yandun2020visual} to determine the equations of motion. We adopt this tree modeling approach to simulate data of the robot interacting with the tree. With this dataset, we train a \textit{forward model} which approximates the resulting state of a deformed tree given its initial state and applied action, as well as a \textit{contact policy} which generates a set of candidate actions to be applied given the tree's initial state and target state.  
A key aspect of our approach is representing the physical structure of a tree as well as its kinematic and dynamic parameters as a graph. This approach is particularly intuitive as it involves a straightforward conversion of the tree geometry in the form of a Euclidean graph. Moreover, this enables the usage of Graph Neural Networks (GNNs), known for their inherent inductive bias properties that allow them to learn latent relationships between nodes and edges in a graph.
This is especially beneficial for tasks like ours, where understanding the interaction between different components in the tree's structure is crucial for accurate modeling and prediction. 

The goal of this study is to establish a framework for modeling trees in the context of learning crop manipulation. Through this line of research, we aim to contribute towards agricultural task automation that demands precise contact interaction with crops. The key contributions of this paper are:
\begin{itemize}
    \item Novel representation of tree crops as graphs, facilitating a GNN-based tree model trained in simulation using mass-spring-damper tree models.
    \item GNN-based policy that infers actions for manipulating trees towards target states while maintaining adaptability to previously unseen tree structures.
    \item Validation of models and policies in a series of experiments and ablation studies conducted in a simulation environment.
\end{itemize}

\section{Related Work}
\label{sec:related_work}
There are several previous studies on automating agricultural tasks with varying degrees of interactions with crops. For instance, \cite{you2022autonomous, silwal2112bumblebee, you2022precision} automates pruning tasks by treating branches as rigid collisions and avoids interacting with crops except with the cutting tool.     
Yandun et al. \cite{yandun2021reaching } leverages reinforcement learning by penalizing collisions with grapevines to reach pruning locations.
Drawing inspiration from how humans use both arms to unclutter crops from leaves, dual-arm robotic systems have been proposed for crop harvesting \cite{sepulveda2020robotic, stavridis2022bimanual, zhang2023push}. As leaves are soft, this degree of interaction can be heuristically addressed \cite{xiong2020autonomous} without explicitly modeling interaction dynamics to plan robot actions. In our work, we propose a general method aimed at intentionally manipulating tree-like crops, with the goal of automating agricultural tasks.

To automate tasks that involve physical interactions with crops, it is essential to understand and model its dynamics. 3D reconstructed crops are often used to reason about their responses to external influences \cite{yandun2020visual, 10160650, katyara2021reproducible}. One common approach to modeling tree dynamics is applying equations of motions based on the spring-damper system \cite{quigley2017real}.  Yandun et al. \cite{yandun2020visual} modeled joint-connected branches as a spring-damper system to predict deformation of trees subjected to external forces. Spatz and Theckes \cite{spatz2013oscillation} studied the damped oscillating behavior of trees when blown by wind. Jacob et al. \cite{jacob2023learning} also modeled trees in simulation as a mass-spring-damper system to estimate system parameters from branch trajectories obtained through active probing.
Our work uses the same simulation setup proposed in \cite{jacob2023learning}, however, our framework leverages graph representations to learn forward models and manipulation policies.


We use graph neural networks to learn the dynamical behavior of tree-like crops.
GNNs are a class of neural networks designed to operate on graphs, enabling them to propagate and aggregate information between nodes and edges to capture latent relationships \cite{scarselli2008graph, li2015gated, gilmer2017neural}. Building upon this foundation, Battaglia et al. \cite{battaglia2016interaction} demonstrated that GNN models are capable of reasoning about how objects interact within complex systems. They achieve this by making dynamical predictions and inferences regarding system parameters. Furthermore, Sanchez-Gonzalez et al. \cite{sanchez2020learning} extended the application of GNNs to simulate complex particle physics, and proposed the use of GNNs as learnable physics engines for dynamical systems \cite{sanchez2018graph}. In our work, we build upon the framework introduced by Sanchez-Gonzalez et al. by utilizing their graph2graph GNN layer \cite{sanchez2018graph} as the backbone of our model. 
This allows us to effectively capture the steady-state behavior of tree-like crops, as well as learning a policy for contact manipulation. 

\section{Methodology}
\label{sec:methodology}
The following section describes our simulation environment and our approach to learning the forward model to predict how trees deform and the contact policy for manipulating tree-like crops.

\subsection{Simulation Environment}
\label{subsec:simulation_environment}

\begin{figure}[H]
\centering 
\includegraphics[width=\columnwidth]{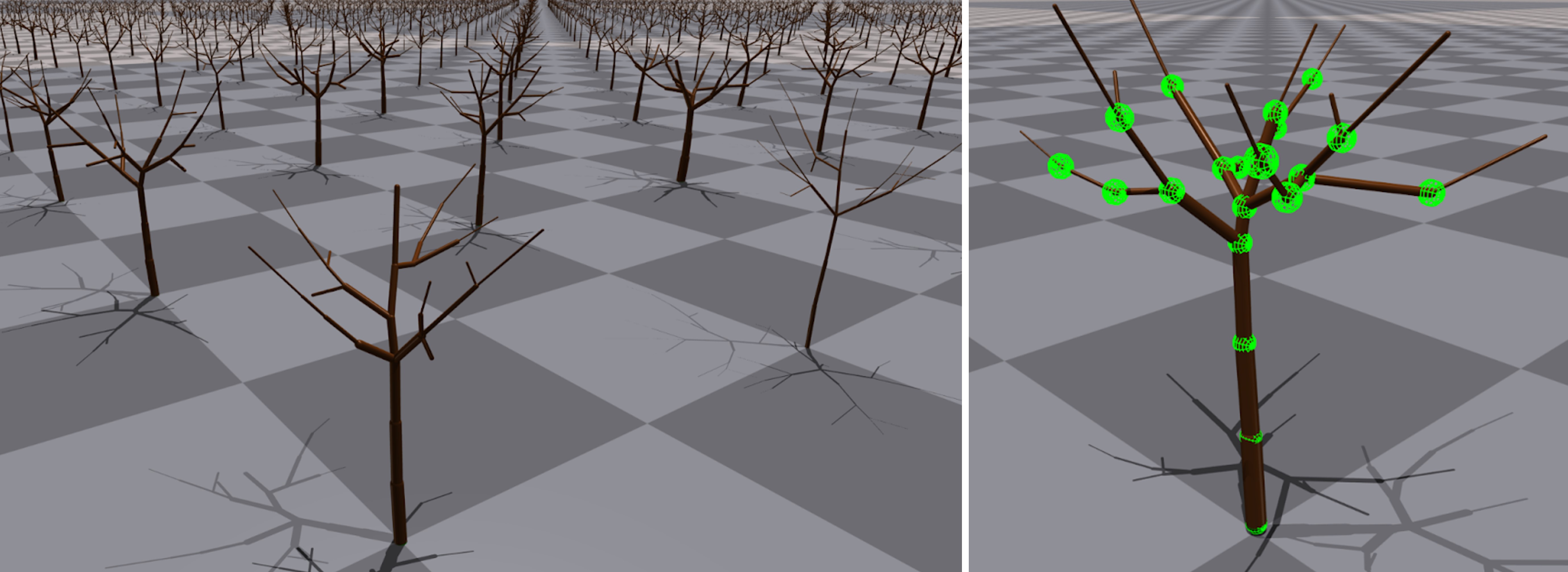}
\vspace*{-3mm}
\setlength{\unitlength}{1cm}
\begin{picture}(0,0)
\put(-1.8, 0.15){\footnotesize (a)}
\put(2.5,  0.15){\footnotesize (b)} 
\put(3.3,  2.2){\textcolor{green}{\footnotesize \textbf{Joint}}} 
\put(2.8,  1.35){\textcolor{brown}{\footnotesize \textbf{Link}}} 
\end{picture}
\vspace{2pt}
\caption{Example of our tree crop model visualized in simulation. (a)  Varying tree sizes and structures are generated in a parallelized environment. (b) The tree is modeled as a series of
rigid links (brown) articulated by spring-damper joints (green).}
\label{fig:simulation_environment} 
\vspace{-10pt}
\end{figure}
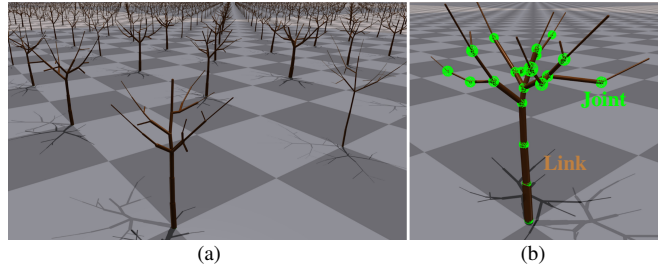

We utilize NVIDIA Isaac Gym \cite{makoviychuk2021isaac} to simulate deformation of multiple tree structures concurrently as visualized in Fig.~\ref{fig:simulation_environment}(a). 
To generate tree structures in simulation that reflect physical tree growth patterns, we use the Space Colonization Algorithm (SCA) as proposed in \cite{runions2007modeling}. The SCA randomly distributes attraction points while iteratively linking tree skeletons to emulate the competition for space between branches that naturally occur in trees. 
To simulate realistic tree movements, we assume that the kinematics and dynamics of trees can be approximated through a series of rigid links as in \cite{yandun2020visual}. These links are connected by spherical joints characterized as spring-damper systems. Specifically, branches are represented as cylindrical links, and spherical joints interconnect two or more branches within the tree's structure (see Fig.~\ref{fig:simulation_environment}(b)). The motion of branches is simulated using a proportional-derivative (PD) controller, governed by adjustable stiffness and damping parameters based on the spring-damper model:
\begin{equation} \label{eq:1}
    \tau = K_s\theta_p + K_d\theta_v
\end{equation} 
where \( \tau \) represents the applied torque, \( K_s \) and \( K_d \) are the stiffness and damping coefficients respectively, \( \theta_p \) is the angular position error, and \( \theta_v \) is the angular velocity error.

We populate the stiffness coefficients $K_s$ based on the beam deflection model \cite{timoshenko1953history}:
\begin{equation} \label{eq:2}
    \theta=\frac{F\ell^2}{2EI} 
\end{equation}
where \( \theta \) is the angular deflection, \( F \) is the applied force, \( \ell \) is the length of the beam, \( E \) is the elastic modulus of the material, and \( I \) denotes the second moment of area of the cross-section of the beam.
Assuming that a tree with circular branch cross-sections undergoing deformation has reached steady-state, substituting equation (\ref{eq:2}) into equation (\ref{eq:1}) yields:
\begin{equation} \label{eq:3}
    K_s=\frac{E\pi r^4}{2\ell}
\end{equation}
where $r$ and $\ell$ denotes the cross-sectional radius and length of the branch (generated from SCA), respectively.
The damping coefficients are empirically set to be $K_d=K_s/10$
. The topology of the tree, as well as the control ($K_s$, $K_d$) and geometric ($r$, $\ell$) parameters are stored and loaded using the Unified Robotics Description Format.


\subsection{Dataset Collection}
\label{subsec:dataset_collection}

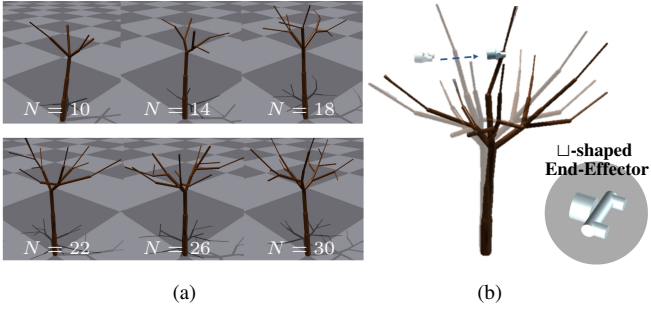
\begin{figure}[H]
\centering 
\includegraphics[width=\columnwidth]{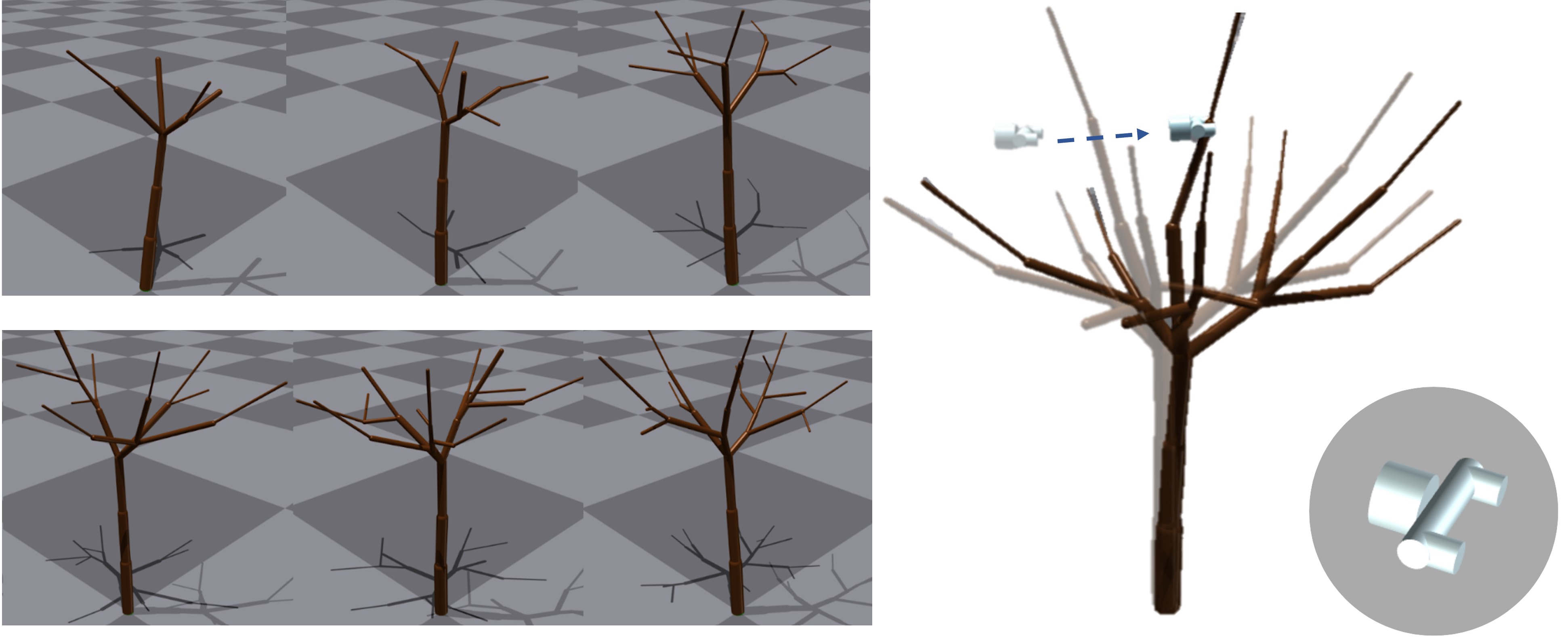}
\setlength{\unitlength}{1cm}
\begin{picture}(0,0)


\put(-4.05, 2.45){\textcolor{white}{\scriptsize $N=10$}} 
\put(-2.45, 2.45){\textcolor{white}{\scriptsize $N=14$}}
\put(-0.8, 2.45){\textcolor{white}{\scriptsize $N=18$}}
\put(-4.05, 0.6){\textcolor{white}{\scriptsize $N=22$}}
\put(-2.45, 0.6){\textcolor{white}{\scriptsize $N=26$}}
\put(-0.8, 0.6){\textcolor{white}{\scriptsize $N=30$}}

\put(3.05, 1.9){\scriptsize $\sqcup$\textbf{-shaped}} 
\put(2.9, 1.65){\scriptsize \textbf{End-Effector}} 

\put(-2.05, 0.0){\footnotesize (a)} 
\put(2.0,  0.0){\footnotesize (b)} 

\end{picture}
\caption{Visualization of the dataset collection. (a) Trees with varying sizes and randomized structures are generated as discussed in Sec. \ref{subsec:simulation_environment}. (b) A $\sqcup$-shaped end-effector makes contact and pushes the tree by tracking a linear trajectory. The initial tree state, end-effector trajectory, contact node, and the final tree state is recorded for each push interaction episode.}
\label{fig:data_collection} 
\vspace{-13pt}
\end{figure}

We abstract the physical structure of a tree as a directed graph denoted by $G=(\mathbf{g}, \{\mathbf{n}_j\}_{j=1\cdots N},\{\mathbf{e}_{ij}\})$, where $\mathbf{g}$ is a vector of global attributes,  $\{\mathbf{n}_j\}_{j=1\cdots N}$ consists of $N$ nodes representing branch locations in the tree with node attributes $\mathbf{n}_j$, and $\{\mathbf{e}_{ij}\}$ consists of edges representing cylindrical branches connecting parent node $\mathbf{n}_{i}$ and child node $\mathbf{n}_{j}$ with attributes $\mathbf{e}_{ij}$. 
Using our tree generation pipeline (Sec.~\ref{subsec:simulation_environment}), we generate 100 unique tree topologies for each tree size ranging from 10 to 30 nodes (with increments of 1) as illustrated in Fig.~\ref{fig:data_collection}(a). The height of the trees range between 0.6m and 1.7m. In order to simulate and acquire data pertaining to tree contact manipulation, we employ an unconstrained $\sqcup$-shaped rigid end-effector that pushes the tree. The end-effector initiates contact with a tree node and subsequently follows a linear trajectory along the direction of the wrist as shown in Fig.~\ref{fig:data_collection}(b). The selection of the contact node, as well as the direction and distance of the end-effector trajectory are sampled from a uniform distribution. 

Each data point is composed from a push interaction episode by recording the initial node absolute positions $\mathbf{p}_j=[p_x, p_y, p_z]$, displaced final node absolute positions $\mathbf{p}'_j=[p'_x, p'_y, p'_z]$, end-effector trajectory $\mathcal{T}=[dx, dy, dz]$, contact node flag  $c_j$, and stiffness coefficients of joints $K_s$. The contact node flag indicates which tree node the end-effector makes contact with ($c_j=1$ if node is in contact, 0 otherwise), hence only one node is set as the contact node in each episode. We collected a training dataset comprising 6,000 push interactions for each tree size category with node counts of $N\in\{11, 13, 15, 17, 19\}$, and a testing dataset comprising 1,000 push interactions for each tree size category spanning the range of $N\in[10,30]$, resulting in a combined dataset of 51,000 push interactions.


\subsection{Learning the Forward Model \& Contact Policy}

\subsubsection{Forward Model}
The forward model problem involves predicting the tree's future state, based on its initial state and the actions performed by the end-effector (see Fig.~\ref{fig:input_output}(a)). We encode the input graph $G_\text{in}$ with the following global, node, and edge attributes:
\begin{align}
    \mathbf{g}&=[\mathcal{T}]\label{eq:4}\\
    \mathbf{n}_j&=[c_j] \label{eq:5}\\
    \mathbf{e}_{ij}&=[\mathbf{p}_j-\mathbf{p}_i, v_{ij}, K_s] \label{eq:6}
\end{align}
where 
$v_{ij}=1$ if the direction of the edge is along the direction from the tree's root-to-leaf, or $-1$ otherwise. 
The forward model is trained to predict the positional differences $\Delta \mathbf{p}_j = [\Delta p_x, \Delta p_y, \Delta p_z]$ for each node at steady-state. Hence, the absolute node positions are obtained by updating the initial node positions with the predicted differences.

\subsubsection{Contact Policy}
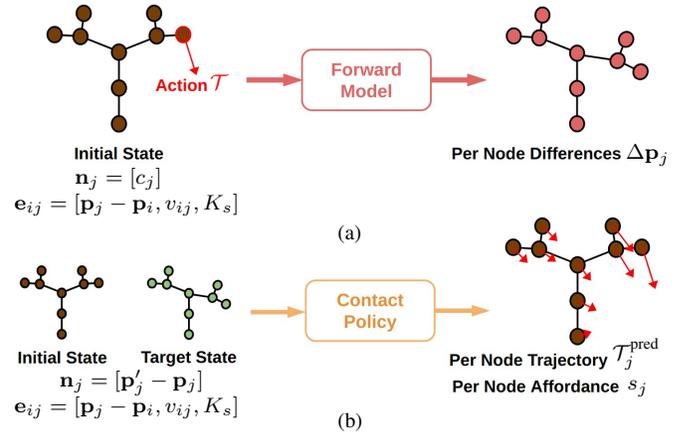
\begin{figure}[H]
\centering 
\includegraphics[width=\columnwidth]{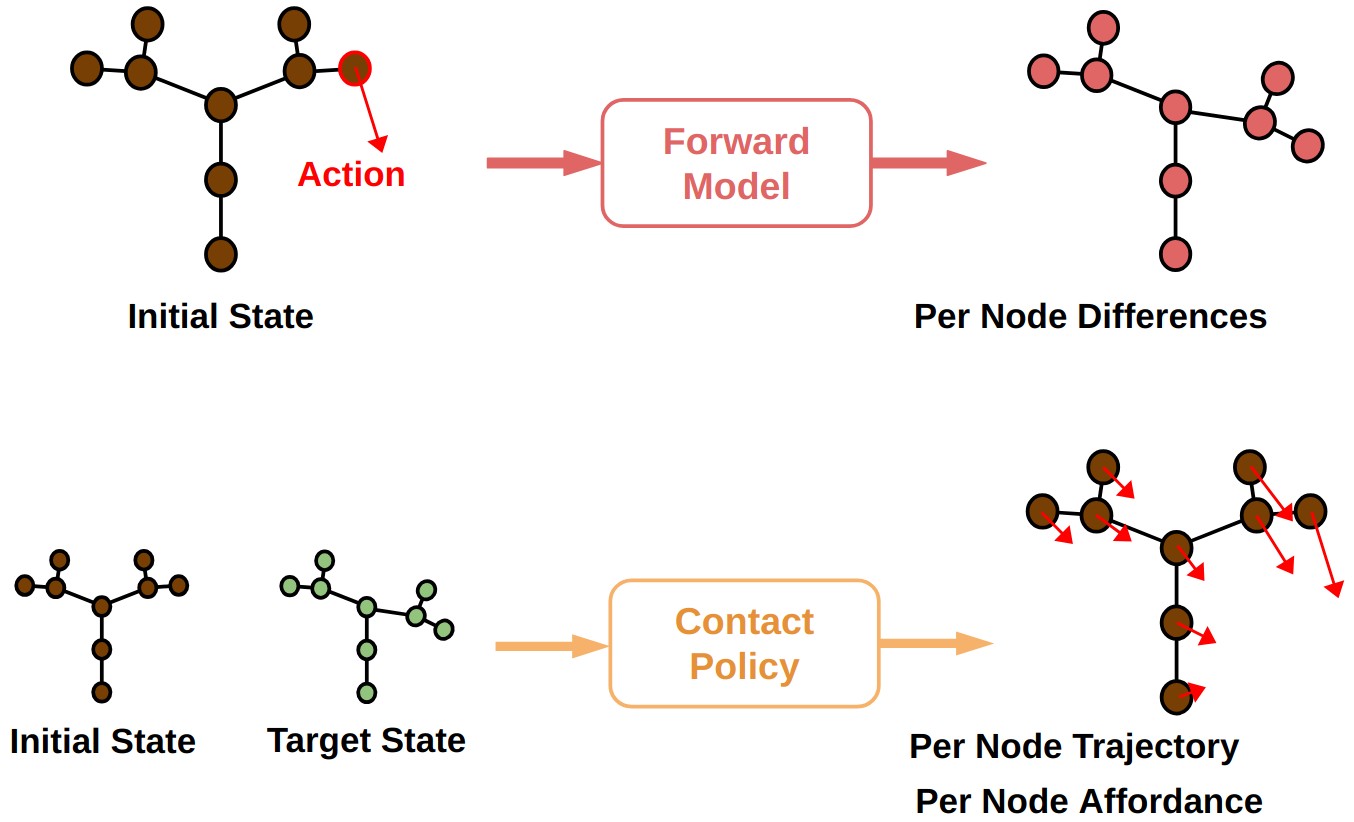}
\setlength{\unitlength}{1cm}
\begin{picture}(0,0)
\put( 0.0, 2.5){\footnotesize (a)} 
\put( -1.7, 4.5){\footnotesize \textcolor{red}{$\mathcal{T}$}} 
\put( -3.5, 3.25){\footnotesize $\mathbf{n}_j=[c_j]$} 
\put( -4.3, 2.9){\footnotesize $\mathbf{e}_{ij}=[\mathbf{p}_j-\mathbf{p}_i, v_{ij}, K_s]$} 
\put( 3.83, 3.57){\footnotesize $\Delta \mathbf{p}_j$}

\put(0.0,  0.0){\footnotesize (b)} 
\put( -3.7, 0.55){\footnotesize $\mathbf{n}_j=[\mathbf{p}'_j-\mathbf{p}_j]$} 
\put( -4.3, 0.2){\footnotesize $\mathbf{e}_{ij}=[\mathbf{p}_j-\mathbf{p}_i, v_{ij}, K_s]$} 
\put( 3.65, 0.9){\footnotesize $\mathcal{T}_j^\text{pred}$} 
\put( 3.85, 0.5){\footnotesize $s_j$} 






\end{picture}
\caption{The input and output diagrams of the (a) forward model and (b) contact policy.}
\label{fig:input_output} 
\vspace{-10pt}
\end{figure}

The contact policy problem is to predict the contact location and action required from the end-effector, given information on the trees current state and its desired target state (see Fig.~\ref{fig:input_output}(b)). The input graph $G_\text{in}$ for the contact policy shares the same edge attributes as the forward model in equation~(\ref{eq:6}), however, the global attributes is null and the node attributes is defined as:
\begin{equation}
    \mathbf{n}_j=[\mathbf{p}'_j-\mathbf{p}_j]
\end{equation}
which encodes per node position difference between the target state and current state. The target states are set to be the final node positions recorded in simulation which ensures feasibility.
We train our contact policy to predict a \textit{per-node} linear end-effector trajectory $\mathcal{T}_j^\text{pred}=[dx, dy, dz]$, and a \textit{per-node} affordance score $s_j$. 
The affordance score ranges between 0 and 1 such that $\sum_{j=1}^Ns_j=1$,  quantifying which node the end-effector should make contact with relative to all other nodes in the tree. 








\begin{figure*}[h!]
\centering 
\includegraphics[width=\textwidth]{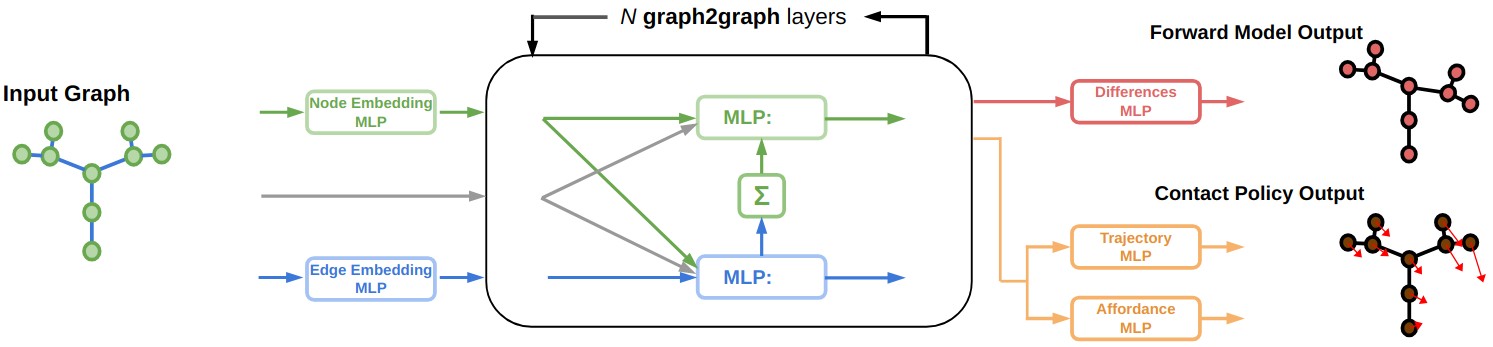}
\setlength{\unitlength}{1cm}
\begin{picture}(0,0)
\put(-7.1, 3.35){\footnotesize $G_\text{in}$} 

\put(-6.4, 3.1){\footnotesize \node{$\{\mathbf{n}_j\}$}} 
\put(-6.1, 2.15){\footnotesize $\mathbf{g}$} 
\put(-6.45, 1.15){\footnotesize \edge{$\{\mathbf{e}_{ij}\}$}}

\put(-3.0, 3.1){\footnotesize \node{$\{\mathbf{n}_j\}$}} 
\put(-2.8, 2.15){\footnotesize $\mathbf{g}$} 
\put(-3.0, 1.15){\footnotesize \edge{$\{\mathbf{e}_{ij}\}$}} 

\put(0.5, 3.1){\footnotesize \node{$f_n$}} 
\put(0.5, 1.2){\footnotesize \edge{$f_e$}} 

\put(0.4, 2.6){\scriptsize \node{$\{\hat{\mathbf{e}}_{j}\}$}} 
\put(0.4, 1.65){\scriptsize \edge{$\{\mathbf{e}^*_{ij}\}$}}

\put(2.0, 3.1){\footnotesize \node{$\{\mathbf{n}^*_j\}$}} 
\put(2.0, 1.15){\footnotesize \edge{$\{\mathbf{e}^*_{ij}\}$}} 

\put(6.1, 3.27){\footnotesize $\{\Delta \mathbf{p}_j\}$}
\put(6.1, 1.53){\footnotesize $\{\mathcal{T}^\text{pred}_j\}$} 
\put(6.2, 0.7){\footnotesize $\{s_{j}\}$}

\end{picture}
\vspace{-15pt}
\caption{Our GNN model architecture utilizes stackable graph2graph layers \cite{battaglia2016interaction}, placed between input embedding layers and output prediction heads. The model takes as input a graph $G_\text{in}$ and outputs per-node predictions. Different output heads are used for the forward model and the contact policy.}
\label{fig:model_architecture} 
\vspace{-18pt}
\end{figure*}
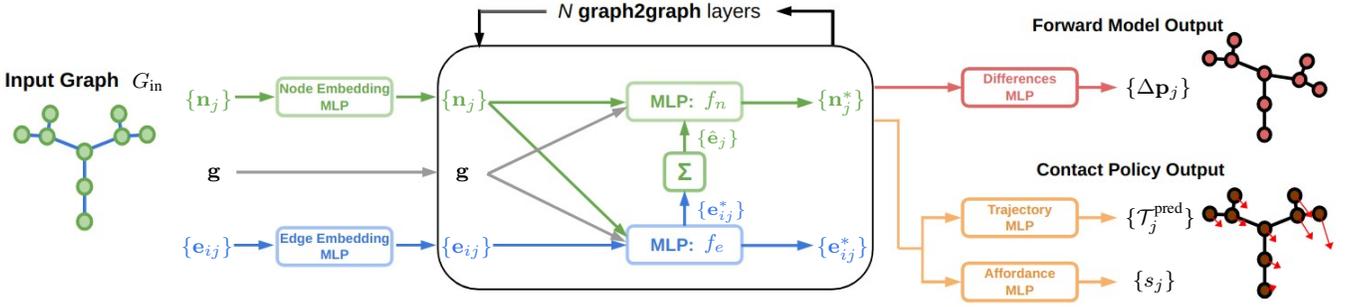

\subsection{GNN Model}

We use the graph2graph layer proposed in \cite{sanchez2018graph} for our GNN model to learn both the forward model as well as the contact policy. The graph2graph layer consists of two sub-functions, $f_e$ and $f_n$ implemented as multilayer perceptrons (MLP), to compute edge and node feature embeddings as follows:
\begin{enumerate}
    \item For each edge $\mathbf{e}_{ij}$, compute new edge feature embedding $\mathbf{e}^*_{ij}$: 
    \begin{equation}
        \mathbf{e}^*_{ij}=f_e(\mathbf{g}, \mathbf{n}_i, \mathbf{n}_j, \mathbf{e}_{ij})
    \end{equation}
    \item For each node $\mathbf{n}_j$, aggregate incoming edge embeddings: 
    \begin{equation}
        \hat{\mathbf{e}}_{j}=\sum_i \mathbf{e}^*_{ij}
    \end{equation} 
    and subsequently compute new node embeddings $\mathbf{n}^*_j$: 
    \begin{equation}
        \mathbf{n}^*_j=f_n(\mathbf{g}, \mathbf{n}_j, \hat{\mathbf{e}}_j)
    \end{equation}
\end{enumerate}
The graph2graph layer is stacked such that the output node and edge embedding of the current layer is passed as input to the next layer. We empirically set our GNN model to contain five graph2graph layers placed between preprocessing MLPs that extract node/edge feature's from their raw attributes, and post-processing MLP output heads. The overall architecture of our GNN model is shown in Fig.~\ref{fig:model_architecture}, where 
the forward model and the contact policy shares the same GNN backbone, and differs only by their input attributes and output heads.

We further note that the input graph is preprocessed into a fully connected graph (as opposed to having a partially connected graph with edges only at physical branches) prior to being passed into the model.
Information of the tree structure is preserved by setting $K_s=0$ for non-branch edges, while $v_{ij}$ preserves the tree growth direction from root-to-leaf.
The reason for this is to facilitate efficient communication between all nodes and edges. Preliminary tests suggested this provides significant performance advantages over a partially connected graph, further discussed in our ablation studies in Sec.~\ref{subsec:ablation_study}. 












The GNN model is implemented using Pytorch Geometric \cite{Fey/Lenssen/2019}. All MLPs in Fig.~\ref{fig:model_architecture} are composed of three fully connected layers with a hidden-size of 128. The forward model and contact policy are trained separately, hence the two models do not share weights. We use Adam with a batch size of 64 and learning rate of 0.001 to optimize the model.

\subsection{Contact Policy with a Robot Arm}
\label{subsec:planning_algorithm}

Given an initial state and target state of a tree, the contact policy outputs a set of candidate actions represented as per-node trajectories
$\{\mathcal{T}_i^\text{pred}\}$ and affordance scores $\{s_i\}$. The contact policy, however, cannot be directly executed by a robot arm because the training data is collected using a free-floating end-effector. 
In order to address the robot's joint configuration space, workspace limitations and collision constraints that may arise from the disparity between a free-floating end-effector and a robot arm, we check the feasibility of candidate actions and select the best action. We achieve this by making use of the multimodal solution offered by the per-node trajectories $\{\mathcal{T}_i^\text{pred}\}$ and affordance scores $\{s_i\}$ as presented in Algorithm 1. We iteratively apply the RRT* algorithm to each node, prioritized by decreasing affordance scores, aiming to identify a valid joint space trajectory for a robot arm to reach and interact with the candidate node. If a feasible path is identified, the policy can be executed. Otherwise, we proceed to the next candidate node and repeat the process.

\vspace{-3pt}
\begin{table}[h]
\centering
\small
\begin{tabularx}{\columnwidth}{l}
\toprule
\makecell[cl]{\textbf{Algorithm 1} Contact Policy Trajectory Planner} \\ 
\midrule
\makecell[cl]{
\textbf{Input:} Input Graph $G_\text{in}$\\
\textbf{Output:} Joint Space Trajectory $\mathcal{J}$\\
{\footnotesize 1:} \hspace{1mm} $\{\mathcal{T}_i^\text{pred}\}$, $\{s_i\}$ $\leftarrow\texttt{ContactPolicy}(G_\text{in})$\\
{\footnotesize 2:} \hspace{1mm} $i_\text{order}\leftarrow\texttt{ArgSortDescending}(\{s_i\})$\\
{\footnotesize 3:} \hspace{1mm} \textbf{for} $i$ \textbf{in} $i_\text{order}$\\
{\footnotesize 4:} \hspace{5mm} $\mathcal{J}\leftarrow\texttt{RRTStar}(\mathcal{T}_i^\text{pred})$\\
{\footnotesize 5:} \hspace{5mm} \textbf{if} $\mathcal{J}$ is feasible\\
{\footnotesize 6:} \hspace{9mm} \textbf{return} $\mathcal{J}$\\
} \\ 
\bottomrule
\end{tabularx}
\vspace{-13pt}
\end{table}

\section{Experiment Results}
\label{sec:experiments}
\subsection{Metric \& Baselines}

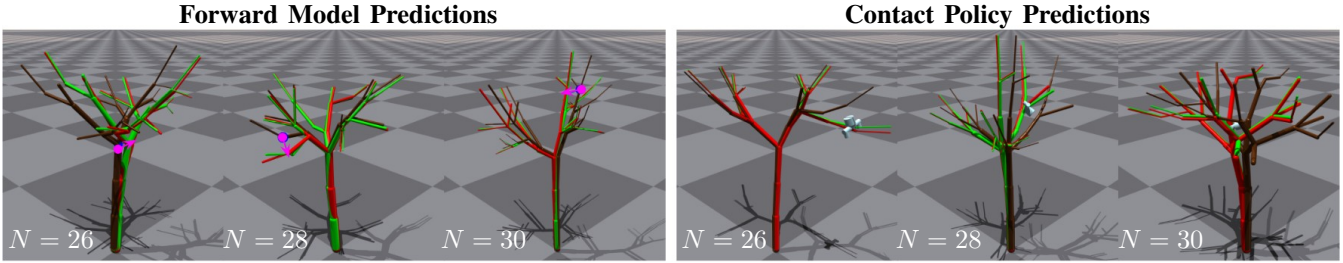
\begin{figure*}[h!]
\centering 
\includegraphics[width=\textwidth]{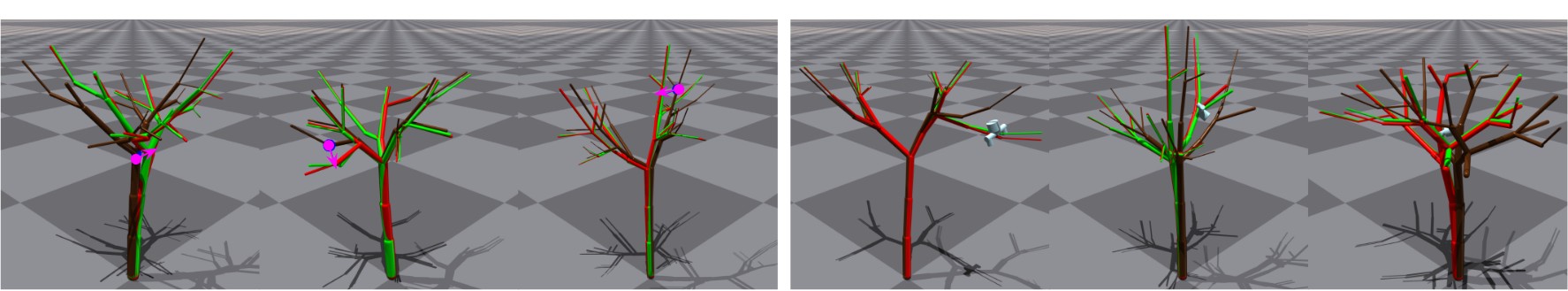}
\setlength{\unitlength}{1cm}
\begin{picture}(0,0)
\put(-6.6, 3.6){ \textbf{Forward Model Predictions}} 
\put(2.25, 3.6){ \textbf{Contact Policy Predictions}} 

\put(-8.75, 0.6){\textcolor{white}{$N=26$}} 
\put(-5.9, 0.6){\textcolor{white}{$N=28$}} 
\put(-3.0, 0.6){\textcolor{white}{$N=30$}} 

\put(0.2, 0.6){\textcolor{white}{$N=26$}} 
\put(3.05, 0.6){\textcolor{white}{$N=28$}} 
\put(6.0, 0.6){\textcolor{white}{$N=30$}}

\end{picture}
\vspace{-13pt}
\caption{Visualization of forward model and contact policy predictions on tree sizes of $N\in\{26, 28, 30\}$. The initial tree, ground truth target tree, and predicted tree is shown in the colors brown, green, and red, respectively. The magenta arrow in the forward model predictions depict the applied action. }
\label{fig:results_visualization} 
\vspace{-8pt}
\end{figure*}

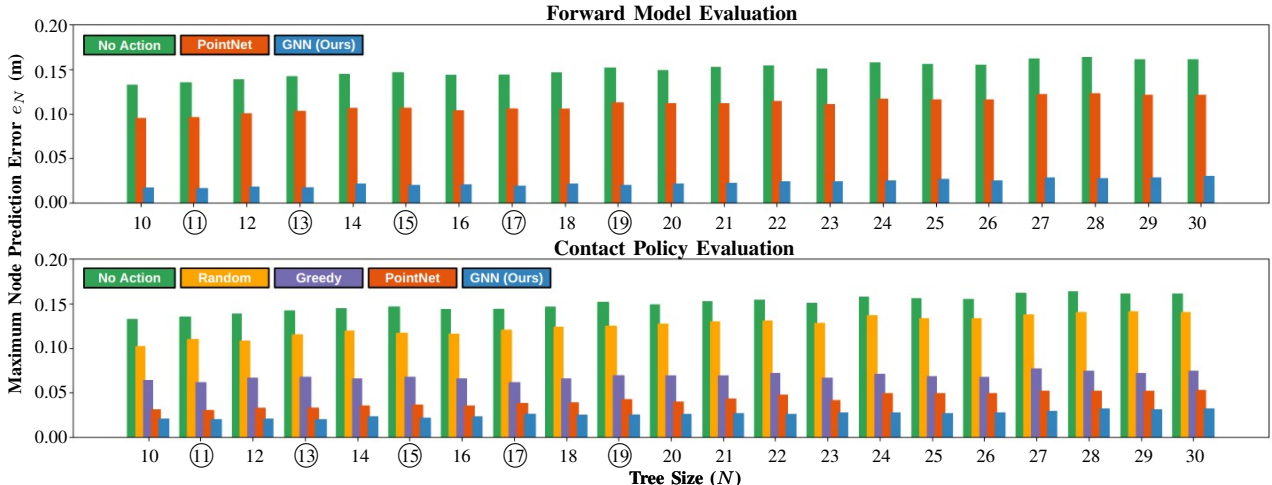
\begin{figure*}[h!]
\centering 
\includegraphics[width=0.9\textwidth]{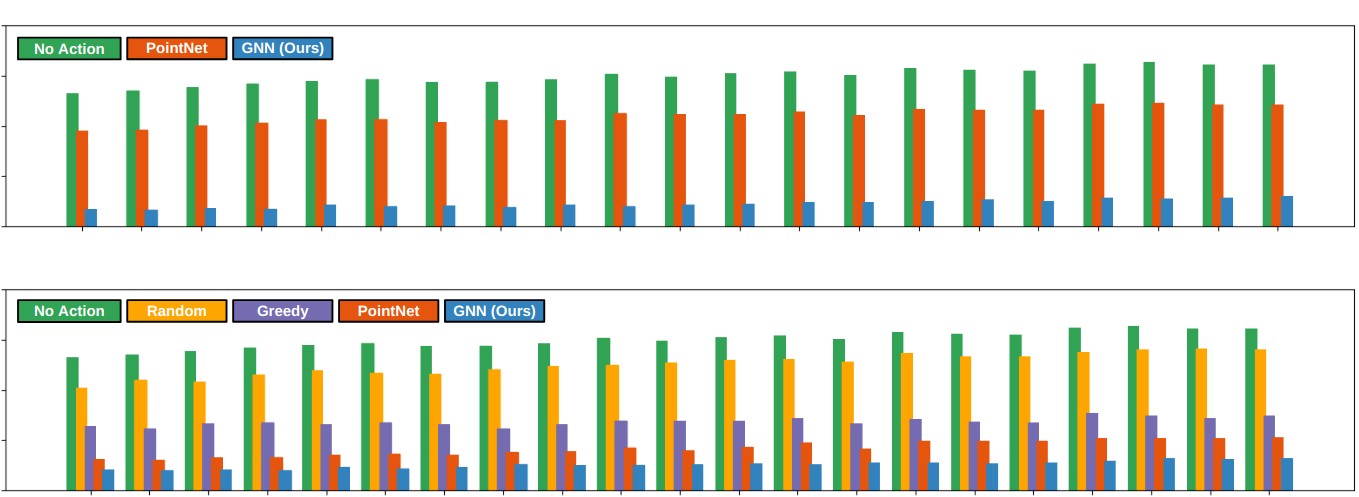}
\setlength{\unitlength}{1cm}
\begin{picture}(-1,0)

\put(-9.9, 5.62){\footnotesize \textbf{Forward Model Evaluation}} 
\put(-9.8, 2.5){\footnotesize \textbf{Contact Policy Evaluation}} 

\put(-17.05, 0.65){\scriptsize \rotatebox{90}{\textbf{Maximum Node Prediction Error $e_{N}$ (m)}}} 
\put(-16.7, 3.12){\scriptsize 0.00} 
\put(-16.7, 3.72){\scriptsize 0.05} 
\put(-16.7, 4.3){\scriptsize 0.10} 
\put(-16.7, 4.86){\scriptsize 0.15} 
\put(-16.7, 5.46){\scriptsize 0.20} 

\put(-8.8, -0.55){\scriptsize \textbf{Tree Size ($N$)}} 
\put(-15.4, 2.85){\scriptsize 10} 
\put(-14.87, 2.85){ \textcircled{\raisebox{-0.5pt}{\scriptsize 11}}} 
\put(-14.0, 2.85){\scriptsize 12} 
\put(-13.47, 2.85){ \textcircled{\raisebox{-0.5pt}{\scriptsize 13}}} 
\put(-12.6, 2.85){\scriptsize 14} 
\put(-12.07, 2.85){ \textcircled{\raisebox{-0.5pt}{\scriptsize 15}}} 
\put(-11.17, 2.85){\scriptsize 16} 
\put(-10.63, 2.85){ \textcircled{\raisebox{-0.5pt}{\scriptsize 17}}} 
\put(-9.75, 2.85){\scriptsize 18} 
\put(-9.22, 2.85){ \textcircled{\raisebox{-0.5pt}{\scriptsize 19}}} 
\put(-8.35, 2.85){\scriptsize 20} 
\put(-7.65, 2.85){\scriptsize 21} 
\put(-6.95, 2.85){\scriptsize 22} 
\put(-6.25, 2.85){\scriptsize 23} 
\put(-5.55, 2.85){\scriptsize 24} 
\put(-4.85, 2.85){\scriptsize 25} 
\put(-4.15, 2.85){\scriptsize 26} 
\put(-3.45, 2.85){\scriptsize 27} 
\put(-2.72, 2.85){\scriptsize 28} 
\put(-2.02, 2.85){\scriptsize 29} 
\put(-1.37, 2.85){\scriptsize 30} 

\put(-16.7, 0.0){\scriptsize 0.00} 
\put(-16.7, 0.6){\scriptsize 0.05} 
\put(-16.7, 1.2){\scriptsize 0.10} 
\put(-16.7, 1.76){\scriptsize 0.15} 
\put(-16.7, 2.35){\scriptsize 0.20} 

\put(-8.8, -0.55){\scriptsize \textbf{Tree Size ($N$)}} 
\put(-15.3, -0.25){\scriptsize 10} 
\put(-14.78, -0.25){ \textcircled{\raisebox{-0.6pt}{\scriptsize 11}}} 
\put(-13.9, -0.25){\scriptsize 12} 
\put(-13.39, -0.25){ \textcircled{\raisebox{-0.6pt}{\scriptsize 13}}} 
\put(-12.5, -0.25){\scriptsize 14} 
\put(-12.01, -0.25){ \textcircled{\raisebox{-0.6pt}{\scriptsize 15}}} 
\put(-11.15, -0.25){\scriptsize 16} 
\put(-10.61, -0.25){ \textcircled{\raisebox{-0.6pt}{\scriptsize 17}}} 
\put(-9.73, -0.25){\scriptsize 18} 
\put(-9.22, -0.25){ \textcircled{\raisebox{-0.6pt}{\scriptsize 19}}} 
\put(-8.35, -0.25){\scriptsize 20} 
\put(-7.65, -0.25){\scriptsize 21} 
\put(-6.95, -0.25){\scriptsize 22} 
\put(-6.25, -0.25){\scriptsize 23} 
\put(-5.57, -0.25){\scriptsize 24} 
\put(-4.87, -0.25){\scriptsize 25} 
\put(-4.18, -0.25){\scriptsize 26} 
\put(-3.48, -0.25){\scriptsize 27} 
\put(-2.78, -0.25){\scriptsize 28} 
\put(-2.08, -0.25){\scriptsize 29} 
\put(-1.4, -0.25){\scriptsize 30}

\end{picture}
\vspace{13pt}
\caption{Performance of the Forward Model (Top) and Contact Policy (Bottom) against baseline methods. The \textit{No Action} baseline is the error measured when the tree remains still. The bars represent the maximum node prediction error per tree, averaged within the tree size category (refer to equation~(\ref{eq:error})). Circled tree sizes ($N$) were seen during training, while the rest are predicted in a zero-shot manner indicating generality of the model to unseen systems.
}
\label{fig:evaluation_plot} 
\vspace{-15pt}
\end{figure*}

To assess our method, we train and test both the forward model and the contact policy using the dataset described in Sec.~\ref{subsec:dataset_collection}. Our evaluation metric is the error between the predicted node position $\mathbf{p}_j^\text{pred}$ and target node position $\mathbf{p}'_j$ for each tree, averaged within the tree size category $N$:
\begin{equation}\label{eq:error}
    e_{N} = \frac{\sum_{d=1}^D \max_{j\in N}||\mathbf{p}'_j - \mathbf{p}^\text{pred}_j ||}{D}
\end{equation} 
where $D$ is the total number data points (episodes) in category $N$. For each tree, we only measure the maximum node position error to capture the error upper-bound. The forward model directly provides predicted node positions $\mathbf{p}^\text{pred}_j$ by summing the predicted differences to the initial states. However, the contact policy predicts actions instead. Hence, we execute the predicted action in simulation and compare the resulting node positions with the target state for contact policy evaluation. 

For baseline methods, we compare our forward model and contact policy with a learned PointNet \cite{qi2017pointnet} baseline model, which we trained to make predictions on the same data as the GNN model. One important distinction between these two model architectures is that PointNet is specifically designed to handle point cloud data without any edges. Hence, the PointNet model inherently lacks awareness of the underlying tree structure and its associated attributes (recall equation~(\ref{eq:6})). We additionally compare the contact policy with two heuristic baselines: 

\begin{itemize}
    \item \textbf{Greedy Baseline:} The end-effector makes contact with the tree node $\mathbf{n}_j$ that is farthest from its target position, then pushes the tree along the vector $\mathcal{T}=\mathbf{p}'_j-\mathbf{p}_j$.     
    \item \textbf{Random Baseline:} The end-effector makes contact with a randomly chosen node $\mathbf{n}_j$, then pushes the tree along the vector $\mathcal{T}=\mathbf{p}'_j-\mathbf{p}_j$.  
\end{itemize}

\subsection{Evaluation Results}



Using our method, we can accurately replicate the forward tree model with the GNN, and learned policies can push trees to a desired target position. Both forward model and policies generalize to previously unseen trees. Fig.~\ref{fig:results_visualization} illustrates the forward model and contact policy predictions, while Fig.~\ref{fig:evaluation_plot} plots the error metric against tree size category. The GNN-based model outperformed baselines for both the forward model and contact policy predictions. Specifically, our method achieves an average node position error of 2.2cm (GNN) as opposed to 11cm (PointNet) for the forward model predictions, and 2.5cm (GNN) compared to 4.2cm (PointNet) for the contact policy predictions.
We also show that the GNN model can generalize to make zero-shot predictions on more complex tree structures that were held out during training. This is shown in Fig.~\ref{fig:evaluation_plot}, where the model was trained on a training dataset with a maximum tree size of $N=19$, while consistently performing zero-shot predictions even up to trees of $N=30$.

  
  

\begin{figure}[t]
\centering 
\includegraphics[width=0.85\columnwidth]{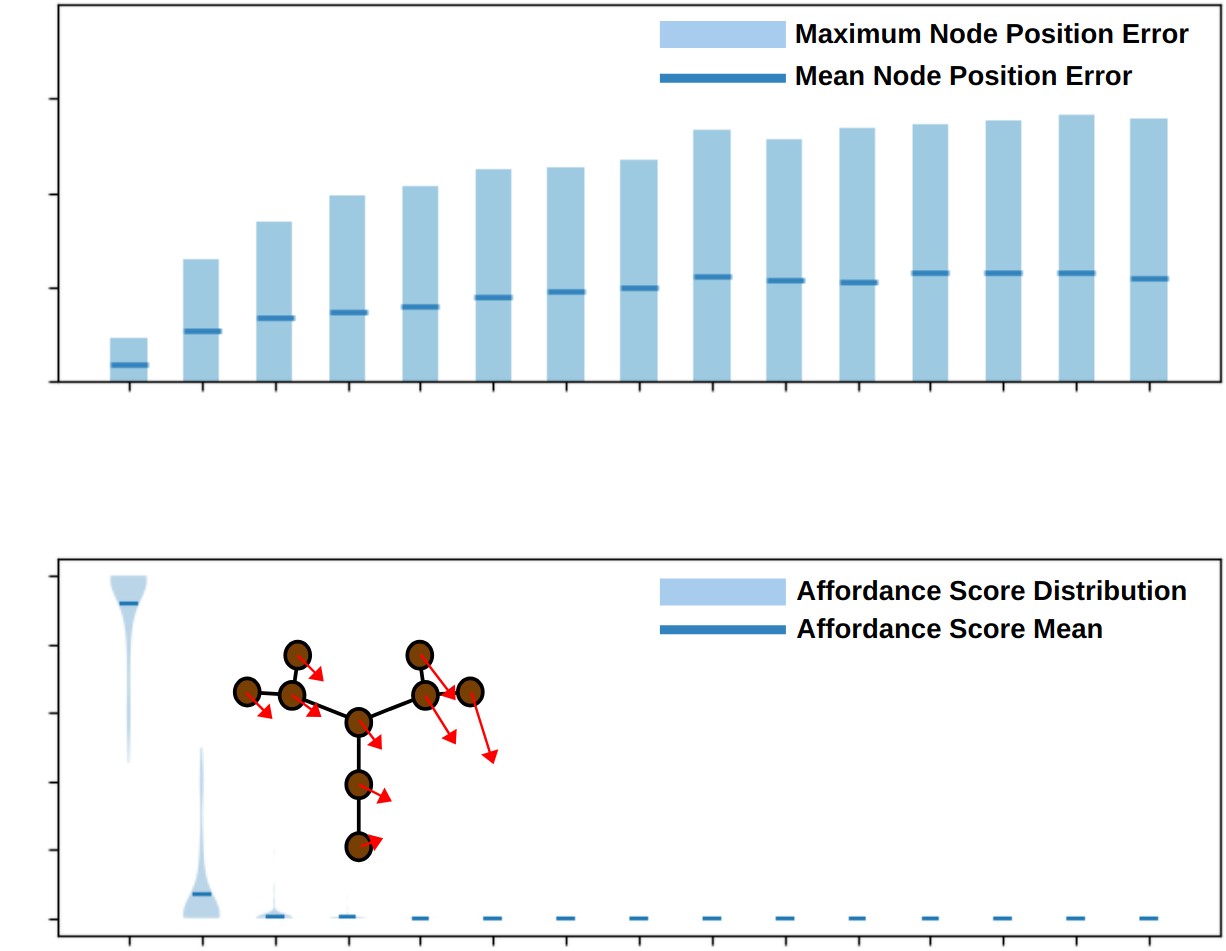}
\setlength{\unitlength}{1cm}
\begin{picture}(-0.5,0)

\put(-3.9, 2.5){\footnotesize (a)} 
\put(-5.0, 2.85){\scriptsize \textbf{Node Affordance Rank}} 

\put(-6.86, 3.13){\scriptsize 1} 
\put(-6.43, 3.13){\scriptsize 2} 
\put(-5.99, 3.13){\scriptsize 3} 
\put(-5.56, 3.13){\scriptsize 4} 
\put(-5.13, 3.13){\scriptsize 5} 
\put(-4.71, 3.13){\scriptsize 6} 
\put(-4.25, 3.13){\scriptsize 7} 
\put(-3.8, 3.13){\scriptsize 8} 
\put(-3.37, 3.13){\scriptsize 9} 
\put(-3.0, 3.13){\scriptsize 10} 
\put(-2.57, 3.13){\scriptsize 11} 
\put(-2.14, 3.13){\scriptsize 12} 
\put(-1.71, 3.13){\scriptsize 13} 
\put(-1.28, 3.13){\scriptsize 14} 
\put(-0.83, 3.13){\scriptsize 15} 

\put(-7.76, 5.03){\scriptsize 0.15} 
\put(-7.76, 4.46){\scriptsize 0.10} 
\put(-7.76, 3.88){\scriptsize 0.05} 
\put(-7.76, 3.33){\scriptsize 0.00}

\put(-8.05, 3.4){\scriptsize \rotatebox{90}{\textbf{Prediction Error (m)}}} 

\put(-5.0, -0.45){\scriptsize \textbf{Node Affordance Rank}} 
\put(-3.9, -0.8){\footnotesize (b)} 

\put(-6.86, -0.17){\scriptsize 1} 
\put(-6.43, -0.17){\scriptsize 2} 
\put(-5.99, -0.17){\scriptsize 3} 
\put(-5.56, -0.17){\scriptsize 4} 
\put(-5.13, -0.17){\scriptsize 5} 
\put(-4.68, -0.17){\scriptsize 6} 
\put(-4.25, -0.17){\scriptsize 7} 
\put(-3.8, -0.17){\scriptsize 8} 
\put(-3.37, -0.17){\scriptsize 9} 
\put(-3.0, -0.17){\scriptsize 10} 
\put(-2.57, -0.17){\scriptsize 11} 
\put(-2.14, -0.17){\scriptsize 12} 
\put(-1.71, -0.17){\scriptsize 13} 
\put(-1.28, -0.17){\scriptsize 14} 
\put(-0.83, -0.17){\scriptsize 15} 

\put(-7.66, 2.15){\scriptsize 1.0} 
\put(-7.66, 1.75){\scriptsize 0.8} 
\put(-7.66, 1.35){\scriptsize 0.6} 
\put(-7.66, 0.95){\scriptsize 0.4} 
\put(-7.66, 0.54){\scriptsize 0.2} 
\put(-7.66, 0.1){\scriptsize 0.0} 

\put(-4.58, 1.47){\scriptsize $s_i=0.7$} 
\put(-4.58, 1.2){\scriptsize $\text{Rank}=1$} 
\put(-5.18, 2.15){\scriptsize $s_j=0.2$} 
\put(-5.18, 1.9){\scriptsize $\text{Rank}=2$}

\put(-7.95, 0.25){\scriptsize \rotatebox{90}{\textbf{Node Affordance}}}

\end{picture}
\vspace{20pt}
\caption{(a) Bar plot showing the prediction error versus the node affordance rank. (b) Violin plot showing the distribution of node affordance scores versus the affordance rank. The embedded tree graph depicts an example of a tree's affordance score and its resulting rank for two of its nodes.}
\label{fig:gnn_node_analysis} 
\vspace{-18pt}
\end{figure}
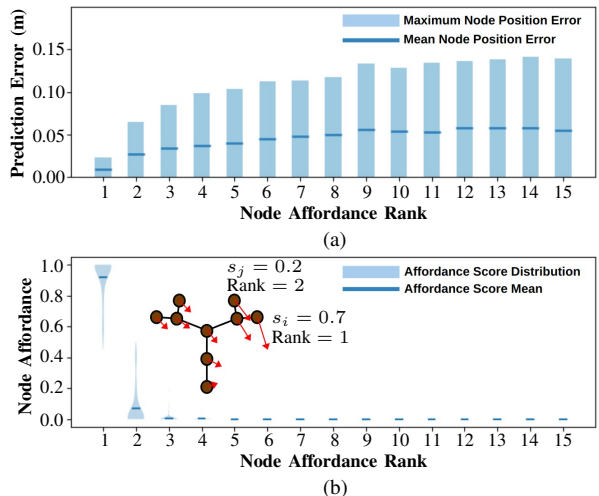

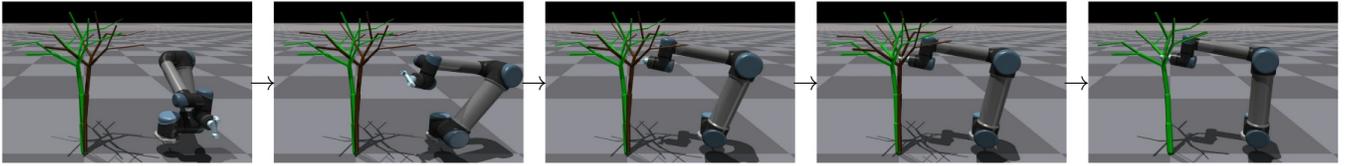
\begin{figure*}[h!]
\centering 
\includegraphics[width=\textwidth]{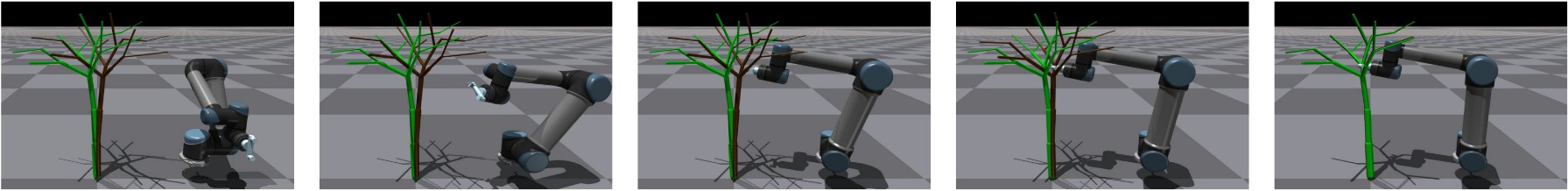}
\setlength{\unitlength}{1cm}
\begin{picture}(0,0)
\put(-5.53, 1.4){$\rightarrow$} 
\put(-1.92, 1.4){$\rightarrow$} 
\put(1.69, 1.4){$\rightarrow$} 
\put(5.29, 1.4){$\rightarrow$} 
\end{picture}
\vspace{-13pt}
\caption{The UR5 robot arm manipulates the tree-crop (brown) to its target state (green) by executing the trajectory obtained from Algorithm 1.}
\label{fig:robot_animation} 
\vspace{-12pt}
\end{figure*}

\subsection{Multimodal Solution with a Simulated Robot Arm}

The affordance scores and actions predicted by the contact policy offer a multimodal solution for manipulating a tree to its target state. We demonstrate this by executing the predicted action for each node and measuring the resulting node position error. 
Fig.~\ref{fig:gnn_node_analysis}(a) illustrates the maximum and average node position errors, while Fig.~\ref{fig:gnn_node_analysis}(b) displays the distribution of affordance scores corresponding to these errors. These metrics are plotted against the node affordance rank (ordered based on decreasing affordance scores), averaged over 100 testing data points for trees with 15 nodes. 
The plot indicates that the position error increases as the affordance score decreases until the sixth highest-scoring node, after which the error tends to plateau. This behavior can be attributed to the tree's branching structure, as contacting one sub-tree versus another may not significantly affect the outcome if neither of them is the primary target for manipulation.   

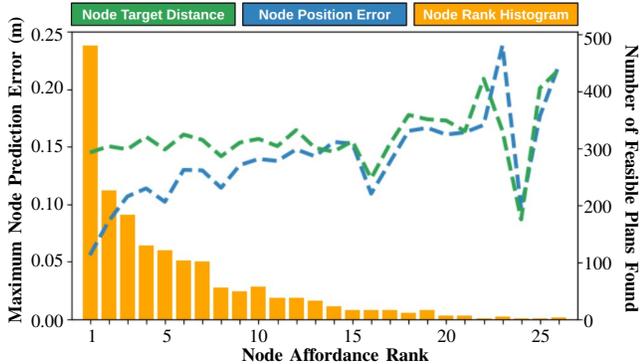
\begin{figure}[t]
\centering 
\includegraphics[width=0.8\columnwidth]{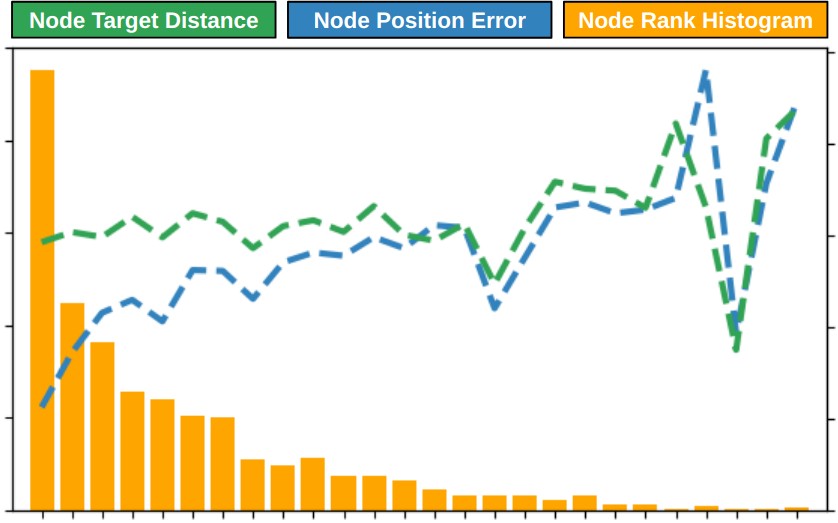}
\setlength{\unitlength}{1cm}
\begin{picture}(-0.3,0)
\put(-4.8, -0.45){\scriptsize \textbf{Node Affordance Rank}} 
\put(-6.85, -0.2){\scriptsize 1} 
\put(-5.85, -0.2){\scriptsize 5} 
\put(-4.7, -0.2){\scriptsize 10} 
\put(-3.45, -0.2){\scriptsize 15} 
\put(-2.2, -0.2){\scriptsize 20} 
\put(-0.95, -0.2){\scriptsize 25} 

\put(-7.9, 0.05){\scriptsize \rotatebox{90}{\textbf{Maximum Node Prediction Error (m)}}} 
\put(-7.6, 0.0){\scriptsize 0.00} 
\put(-7.6, 0.8){\scriptsize 0.05} 
\put(-7.6, 1.55){\scriptsize 0.10} 
\put(-7.6, 2.32){\scriptsize 0.15} 
\put(-7.6, 3.1){\scriptsize 0.20} 
\put(-7.6, 3.8){\scriptsize 0.25} 

\put(0.3, 3.75){\scriptsize \rotatebox{-90}{\textbf{Number of Feasible Plans Found}}} 
\put(-0.25, 0.0){\scriptsize 0} 
\put(-0.25, 0.76){\scriptsize 100} 
\put(-0.25, 1.52){\scriptsize 200} 
\put(-0.25, 2.27){\scriptsize 300} 
\put(-0.25, 3.05){\scriptsize 400} 
\put(-0.25, 3.75){\scriptsize 500}

\end{picture}
\vspace{10pt}
\caption{Evaluating the Contact Policy using a simulated UR5 Robot Arm. The yellow bars display the histogram depicting the number of feasible plans discovered per node affordance rank. A total of 2,100 episodes were tested. The associated initial node target distance (green dashed line) and resulting node position error (blue dashed line) are also plotted.
}
\label{fig:robot_policy_evaluation} 
\vspace{-20pt}
\end{figure}

We apply Algorithm 1 
to execute the actions obtained from the contact policy on a simulated UR5 robot arm as illustrated in Fig.~\ref{fig:robot_animation}. To assess the method, we tested the algorithm on 100 episodes for each tree size ranging from $N\in [10, 30]$, resulting in a planning success rate of $82.9\%$. Fig.~\ref{fig:robot_policy_evaluation} depicts the histogram of successfully planned contact node ranks, along with its corresponding node position errors. The algorithm returned a valid trajectory for the highest ranking node $22.9\%$ of the time, gradually decreasing for lower ranking nodes. The error metric is also consistent with the results obtained in Fig.~\ref{fig:gnn_node_analysis}(a), where the error gradually increases as the node rank increases.


\vspace{-5pt}

\subsection{Ablation Study}
\label{subsec:ablation_study}

\subsubsection{Ablation of Edge Attributes} 
In this study, we aim to show that the GNN model is effectively learning the relationships intrinsic to the tree structure as a graph. To this end, we perform experiments by training the forward model using the same architecture as our proposed method while removing all edge attributes, as well as systematically removing a single edge attribute for each experiment (see Table~\ref{table:ablation_study}, Rows 1-5). The results indicate that the removal of any one edge attribute leads to a higher node prediction error. This underscores the critical role of each edge attribute in enabling the model to accurately predict nodes by learning latent relationships intrinsic to the tree structure.

\subsubsection{Ablation of Input Graph Connectivity} 
We also demonstrate the benefit of preprocessing the input graph into a fully connected form, as opposed to a partially connected one with edges exclusively at physical branches. We train the forward model on partially connected graphs with varying numbers of graph2graph layers, which is then compared to our proposed model trained on fully connected graph inputs  (see Table~\ref{table:ablation_study}, Rows 5-9). Even with an increased number of graph2graph layers, employing a fully connected graph as input for a model with only 5 graph2graph layers (ours) surpasses the performance of models with up to 20 graph2graph layers trained on partially connected input graphs. This indicates that the fully connected graph representation significantly enhances the model's ability to learn and capture latent relationships within the tree graph.

\newcolumntype{A}{>{\centering\arraybackslash\hsize=.2\hsize}X}
\newcolumntype{B}{>{\centering\arraybackslash\hsize=.3\hsize}X}
\newcolumntype{D}{>{\centering\arraybackslash\hsize=.3\hsize}X}
\newcolumntype{E}{>{\centering\arraybackslash\hsize=.5\hsize}X} 
\newcolumntype{F}{>{\centering\arraybackslash\hsize=.3\hsize}X} 
\begin{table}
\caption{Ablation Study of Edge Attributes \& Input Graph Connectivity}
\vspace{-4pt}
\centering 
\begin{tabularx}{\columnwidth}{@{}ABDEF@{}}  
\hline\hline 
\multirow{2}{*}{{\shortstack[c]{\vspace{1pt}\\Ablation \\of}}} & 
\multirow{2}{*}{{\shortstack[c]{\vspace{1pt}\\Number of \\G2G Layers}}} & \multirow{2}{*}{{\shortstack[c]{\vspace{1pt}\\$G_\text{in}$\\Connectivity}}} & \multirow{2}{*}{{\shortstack[c]{\vspace{1pt}\\Included Edge \\Attributes $\mathbf{e}_{ij}$}}} &  \multirow{2}{*}{{\shortstack[c]{\vspace{1pt}\\Prediction\\Error (m)}}} \\[0.5ex] 
& & & \\[0.5ex] 
\hline 
\multirow{4}{*}{{\shortstack[c]{\vspace{1pt}\\\textit{Edge}\\\textit{Attributes}}}} & 5 & Full & $[\cdot]$ & 0.110 \\[0.5ex]
& 5 & Full & $[v_{ij}, K_s]$ & 0.067 \\[0.5ex]
& 5 & Full & $[\mathbf{p}_j-\mathbf{p}_i, v_{ij}]$ & 0.038 \\[0.5ex]
& 5 & Full & $[\mathbf{p}_j-\mathbf{p}_i, K_s]$ & 0.026 \\[0.5ex]
\hdashline
\rule{0pt}{2.5ex}
\textbf{\textit{Ours}} & \textbf{5} & \textbf{Full} & $[\mathbf{p}_j-\mathbf{p}_i, v_{ij}, K_s]$ & \textbf{0.023} \\[0.5ex]
\hdashline
\multirow{4}{*}{{\shortstack[c]{\vspace{1pt}\\\textit{Input}\\\textit{Graph}\\\textit{Connect-}\\\textit{ivity}}}}& 5 & Partial & $[\mathbf{p}_j-\mathbf{p}_i, v_{ij}, K_s]$ & 0.047 \\[0.5ex]
& 10 & Partial & $[\mathbf{p}_j-\mathbf{p}_i, v_{ij}, K_s]$ & 0.039 \\[0.5ex]
& 15 & Partial & $[\mathbf{p}_j-\mathbf{p}_i, v_{ij}, K_s]$ & 0.043 \\[0.5ex]
& 20 & Partial & $[\mathbf{p}_j-\mathbf{p}_i, v_{ij}, K_s]$ & 0.044 \\[0.5ex]
\hline\hline
\end{tabularx}
\label{table:ablation_study} 
\vspace{-18pt}
\end{table}

\section{Conclusion}

In this study, we presented a framework that encodes tree crops, modeled as spring-damper systems, into a graph representation. This enables the learning of both a forward model to predict resulting deformations, and a policy to execute non-prehensile contact actions for tree manipulation, using graph neural networks. Our proposed framework has been comprehensively evaluated in simulation using a simplistic model of tree dynamics. However, validating the method in the real world remains as future work. This involves extensive system identification of the tree's dynamic parameters, as well as consideration of more complex properties (such as the anisotropic characteristics of branches). 
For example, estimating parameters by probing on real trees \cite{jacob2023learning} may lead to realistic simulation for nonlinear dynamical deformation of tree crops.
These steps are imperative to achieve Sim2Real policy transfers for  practical field applications, which we intend to pursue in our next steps. 
Applying our learned forward model to implement model-predictive control or model-based reinforcement learning for task-specific crop manipulation also remains as future work.  


\FloatBarrier

\bibliographystyle{IEEEtran}
\bibliography{references}

\end{document}